\documentclass{article}
\usepackage{ijcai22}
\usepackage[utf8]{inputenc} 
\usepackage[T1]{fontenc}    
\usepackage{hyperref}       
\usepackage{url}            
\usepackage{booktabs}       
\usepackage{amsfonts}       
\usepackage{nicefrac}       
\usepackage{microtype}      

\usepackage{amssymb}
\usepackage{amsmath}
\usepackage{mathtools}
\usepackage{amsfonts}
\usepackage{amsthm,bbm}
\usepackage{color}
\usepackage[para,online,flushleft]{threeparttable}
\usepackage{enumerate}
\usepackage{multirow}
\newtheorem{theorem}{Theorem}

\newtheorem{lemma}{Lemma}
\newtheorem{definition}{Definition}
\theoremstyle{definition}

\newcommand{\argmin}{\mathop{\rm arg~min}\limits}

\newcommand{\Eqref}[1]{Eq.~(\ref{#1})}
\newcommand{\Figref}[1]{Figure~\ref{#1}}
\newcommand{\Tabref}[1]{Table~\ref{#1}}
\newcommand{\Thmref}[1]{Theorem~\ref{#1}}
\newcommand{\Lemref}[1]{Lemma~\ref{#1}}
\usepackage{algorithm}
\usepackage{algorithmic}

\newcommand{\textbfit}[1]{\textbf{\textit{#1}}}

\newcommand{\shortL}{\mathcal{L}^{\ell}}
\newcommand{\idealL}{\mathcal{L}^{\ell}_{ideal}}
\newcommand{\naiveL}{\widehat{\mathcal{L}}^{\ell}_{naive}}
\newcommand{\ipsL}{\widehat{\mathcal{L}}^{\ell}_{IPS}}
\newcommand{\drL}{\widehat{\mathcal{L}}^{\ell}_{DR}}
\newcommand{\causeL}{\widehat{\mathcal{L}}^{\ell}_{CausE}}

\newcommand{\iR}{R_{u,i}}
\newcommand{\ipredR}{\hat{R}_{u,i}}
\newcommand{\trueR}{\boldsymbol{R}}
\newcommand{\predR}{\widehat{\boldsymbol{R}}}
\newcommand{\iO}{O_{u,i}}
\newcommand{\matO}{\boldsymbol{O}}
\newcommand{\iP}{P_{u,i}}
\newcommand{\matP}{\boldsymbol{P}}
\newcommand{\matU}{\boldsymbol{U}}
\newcommand{\matV}{\boldsymbol{V}}

\newcommand{\rad}{\mathfrak{R}_{\boldsymbol{P}, M}}
\newcommand{\prad}{\mathfrak{R}_{\boldsymbol{P}^{\prime}, M}}
\newcommand{\mH}{\mathcal{H}}
\newcommand{\mO}{\mathcal{O}}

\newcommand{\mOmnar}{\boldsymbol{O}_{MNAR}}
\newcommand{\mOmcar}{\boldsymbol{O}_{MCAR}}
\newcommand{\mPmnar}{\boldsymbol{P}_{MNAR}}
\newcommand{\mPmcar}{\boldsymbol{P}_{MCAR}}

\newcommand{\mE}{\mathbb{E}}
\newcommand{\D}{\mathcal{D}}
\newcommand{\SUM}{\sum_{(u,i) \in \D}}

\newcommand{\truePMD}{\psi_{\widehat{\boldsymbol{R}}, \mathcal{H}} \left(\boldsymbol{P}, \boldsymbol{P}^{\prime} \right)}
\newcommand{\shortPMD}{\psi_{\widehat{\boldsymbol{R}}, \mathcal{H}}}

\title{Towards Resolving Propensity Contradiction\\
in Offline Recommender Learning}

\author{
Yuta Saito$^1$
\and
Masahiro Nomura$^2$
\affiliations
$^1$Cornell University,$^2$CyberAgent, Inc.
\emails
ys552@cornell.edu,
nomura\_masahiro@cyberagent.co.jp
}

\begin{document}

\maketitle

\begin{abstract}
We study offline recommender learning from explicit rating feedback in the presence of \textit{selection bias}. A current promising solution for the bias is the \textit{inverse propensity score} (\textsl{IPS}) estimation. However, the performance of existing propensity-based methods can suffer significantly from the propensity estimation bias. In fact, most of the previous IPS-based methods require some amount of \textit{missing-completely-at-random} (\textsl{MCAR}) data to accurately estimate the propensity. This leads to a critical \textbfit{self-contradiction}; IPS is ineffective without MCAR data, even though it originally aims to learn recommenders from only \textit{missing-not-at-random} feedback. To resolve this \textbfit{propensity contradiction}, we derive a \textbfit{propensity-independent generalization error bound} and propose a novel algorithm to minimize the theoretical bound via adversarial learning. Our theory and algorithm do not require a propensity estimation procedure, thereby leading to a well-performing rating predictor without the true propensity information. Extensive experiments demonstrate that the proposed approach is superior to a range of existing methods both in rating prediction and ranking metrics in practical settings without MCAR data.
\end{abstract}

\section{Introduction}
In industrial recommender systems, it is essential to obtain a well-performing rating predictor from sparse rating feedback to recommend items relevant to users.
An important challenge is that most of the missing mechanisms of real-world explicit rating data are \textit{missing-not-at-random} (\textsl{MNAR})~\cite{marlin2009collaborative,schnabel2016recommendations,wang2019doubly,chen2020bias}. 
There are two main causes behind the MNAR mechanism.
The first is the past recommendation policy. 
Suppose that we rely on a policy that recommends popular items with high probabilities.
Then, the observed ratings under that policy include more data of popular items~\cite{bonner2018causal}. 
The second is self-selection by users. 
For example, users tend to rate items for which they exhibit positive preferences, and the ratings for items with negative preferences are more likely to be missing~\cite{marlin2009collaborative,schnabel2016recommendations}. 

\paragraph{\textbf{Open Problem: Propensity Contradiction.}}
Selection bias makes it difficult to learn rating predictors, as naive methods typically result in suboptimal and biased recommendations with MNAR data~\cite{schnabel2016recommendations,steck2010training,wang2019doubly}. 
One of the most established solutions to this problem is the propensity-based approach. 
It defines the probability of each feedback being observed as a \textit{propensity score} and obtains an unbiased estimator for the true metric of interest via inverse propensity weighting~\cite{liang2016causal,schnabel2016recommendations,wang2019doubly}. 
In general, its unbiasedness is desirable, however, this is valid only when the true propensity score is available. 
Previous studies utilize some amount of \textit{missing-completely-at-random} (\textsl{MCAR}) test data to estimate the propensity score and ensure their empirical performance ~\cite{schnabel2016recommendations,wang2019doubly}.
In most real-world recommender systems, however, the true propensity score is unknown, and MCAR data are unavailable as well, resulting in a severe bias in the estimation of the loss function of interest.
Thus, there remains a critical \textbfit{contradiction}: the performance of the IPS variants relies heavily on mostly inaccessible MCAR data, although they originally aim at training recommenders using only MNAR data.

Two previous methods tackle the challenge of those propensity-based methods. 
The first one is \textit{causal embeddings} (CausE)~\cite{bonner2018causal}.
It introduces a domain regularization term to address the bias.
However, CausE requires some amount of MCAR data by its design; it cannot be generalized to a realistic setting with only MNAR training data.
Moreover, its domain regularization technique is a heuristic approach and lacks a theoretical guarantee, and the reason this method works is uncertain.
The other method is to use \textit{1-bit matrix completion} (1BitMC)~\cite{ma2019missing} to estimate propensity scores using only MNAR training data, along with a theoretical guarantee. 
However, the problem is that these methods presuppose the debiasing procedure with \textbf{inverse} propensity weighting and thus cannot be used when there is a user--item pair with zero observed probability. 
Furthermore, the experiments on 1BitMC were conducted using only extremely small datasets (Coat and MovieLens 100k) and prediction accuracy measures (MSE); accordingly, its performance on moderate size benchmark datasets (e.g., Yahoo! R3 \cite{mnih2008probabilistic}) and on a ranking task remain unknown.

\paragraph{\textbf{Contributions.}}
To overcome the limitations of the existing methods, we establish a new theory for offline recommender learning inspired by the theoretical framework of \textit{unsupervised domain adaptation} (UDA)~\cite{ben2010theory}. 
UDA aims to obtain a good predictor in settings where the feature distributions between the training and test sets are different. 
To this end, UDA utilizes distance metrics that measure the dissimilarity between probability distributions and does not depend on the propensity score~\cite{ganin2016domain,saito2017asymmetric,saito2020asymmetric}. 
Thus, the framework is usable even when the true propensity score is unknown and is expected to alleviate the issues caused by the propensity estimation bias in the absence of MCAR data. 
Moreover, the method is valid even when there is a user--item pair with zero observed probability.

To figure out a solution to the propensity contradiction, we first define a novel discrepancy metric to quantify the similarity between two missing mechanisms of rating feedback. Building on our discrepancy measure, we derive a \textbfit{propensity independent generalization error bound} for the loss function of interest.
We further propose \textbfit{domain adversarial matrix factorization}, which minimizes the derived theoretical bound in an adversarial manner.
Our theoretical bound and algorithm are independent of the propensity score and thus address the contradiction of the previous propensity-based solutions.
Finally, we conduct extensive experiments using public real-world datasets. 
In particular, we demonstrate that the proposed approach outperforms existing propensity-based methods in terms of rating prediction and ranking performance under a realistic situation where the true propensity score is inaccessible.

\section{Preliminaries} \label{sec:setup}
Let $u \in [m]$ be a user and $i \in [n]$ be an item in a recommender system. 
We also define $\D := [m] \times [n]$ as the set of all user--item pairs. 
Let $\boldsymbol{R} \in \mathbb{R}^{m \times n}$ denote a (deterministic) true rating matrix, where each entry $\iR$ represents the real-valued true rating of user $u$ for item $i$.

Our goal is to develop an algorithm to obtain a better predicted rating matrix (or a \textit{hypothesis}) $\predR$, where each entry $\ipredR$ denotes a predicted rating value for $(u, i)$.
To this end, we formally define \textbfit{the ideal loss function of interest} that should ideally be optimized to obtain a recommender as follows:
\begin{align}
\idealL ( \predR ) := \frac{1}{mn} \SUM \ell (\iR, \ipredR ), 
\label{eq:ideal_loss}
\end{align}

where $\ell (\cdot, \cdot): \mathbb{R} \times \mathbb{R} \rightarrow [0, \Delta] $ is a loss function bounded by $\Delta$. For example, when $\ell (x, y) = (x - y)^2$, \Eqref{eq:ideal_loss} represents the \textit{mean-squared-error} (MSE).

In real-life recommender systems, one cannot directly calculate the ideal loss function, as most rating data are missing in nature.
To precisely formulate this missing mechanism, we introduce two additional matrices. 
The first one is the \textit{propensity matrix}, denoted as $\matP \in \mathcal{P}$, where $\mathcal{P}$ represents the space of probability distributions over $\D$. 
Each of its entries $\iP \in [0, 1]$ is the \textit{propensity score} of $(u, i)$, and it represents the probability of the rating feedback being observed. 
Next, let $\matO \in \{ 0, 1\}^{m \times n}$ be an \textit{observation matrix} where each entry $\iO \in \{0, 1\}$ is a Bernoulli random variable with its expectation $\mathbb{E}[\iO]=\iP$. 
If $\iO = 1$, the rating of the pair is observed, otherwise, it is unobserved.
Finally, we use $\matO \sim \matP$ when the entries of $\matO$ are the realizations of Bernoulli distributions defined by the entries of $\matP$.
For simplicity and without loss of generality, we assume that $M := \SUM \iO $ hereinafter.

\subsection{Existing Methods}
In our formulation, it is essential to approximate the ideal loss function using only observable feedback.
Here, we summarize the existing methods to estimate the ideal loss function and discuss their limitations.

\paragraph{\textbf{Naive Estimator.}}
Given a set of observed rating feedback, the most basic estimator for the ideal loss is the naive estimator, which is defined as follows:
\begin{align}
\naiveL (\predR; \matO) := \frac{1}{M} \SUM \iO \cdot \ell ( \iR, \ipredR ) . 
\label{eq:naive_loss}
\end{align}

The naive estimator simply averages the loss of user-item pairs with the observed rating feedback (i.e., $O_{u,i}=1$).
This is valid when the missing mechanism is MCAR\footnote{A missing mechanism is said to be MCAR if the propenisty score is constant, i.e., $P_{u,i} = C, \forall(u,i) \in \D$. Ensuring that the data is MCAR is extremely difficult, as the users' self-selection as to which items they provide rating feedback is out of our control.}, as it is unbiased against the ideal loss function with MCAR data~\cite{schnabel2016recommendations}. 
However, several previous studies indicate that this estimator exhibits a bias under general MNAR settings (i.e., $\mE [\naiveL] \neq \idealL$ for some $\predR$). 
This means that the naive estimator does not converge to the ideal loss, even with infinite data.
Therefore, one should use an estimator that addresses the bias as an alternative to the naive one~\cite{schnabel2016recommendations}.

\paragraph{\textbf{Inverse Propensity Score (IPS) Estimator.}}
To improve the naive estimator, several previous studies apply the IPS estimation to alleviate the bias of MNAR rating feedback~\cite{liang2016causal,schnabel2016recommendations,saito2020unbiased,saito2020pairwise,saito2021evaluating}. 
In causal inference, propensity scoring estimators are widely used to estimate the causal effects of a treatment from observational data~\cite{imbens2015causal}. 
In our formulation, we can derive an unbiased estimator for the loss function of interest by using the true propensity score as follows:
\begin{align}
\ipsL (\predR; \matO) := \frac{1}{mn} \SUM \iO \cdot \frac{\ell (\iR, \ipredR )}{\iP} . 
\label{eq:ips_loss}
\end{align}

This estimator is unbiased against the ideal loss (i.e., $ \mE[ \ipsL ] = \idealL $ for any $\predR$) and is more desirable than the naive estimator in terms of bias. 
However, its unbiasedness is ensured only when the true propensity score is available, and it can have a bias with an inaccurate propensity estimator (see Lemma 5.1 of Schnabel et al.~\cite{schnabel2016recommendations}). 
The bias of IPS can arise in most real-world recommender systems, because the missing mechanism of rating feedback can depend on user self-selection, which cannot be controlled by analysts and is difficult to estimate~\cite{marlin2009collaborative,schnabel2016recommendations}. 
Indeed, most previous studies estimate the propensity score by using some amount of MCAR test data to ensure empirical performance~\cite{schnabel2016recommendations,wang2019doubly}. 
However, this is infeasible owing to the costly annotation process~\cite{joachims2017unbiased}. 
In addition, this method is valid only when the ratings for some user--item pairs are observed with non-zero probabilities (i.e., $\iP \in (0,1], \forall (u,i) \in \D$), which is a criterion difficult to verify in practice~\cite{ma2019missing}.
Therefore, we explore theory and algorithm independent of the propensity score and MCAR data, aiming to alleviate the concurrent issues of the propensity-based methods.

\paragraph{\textbf{Doubly Robust (DR) Estimator.}}
The DR estimator utilizes the error imputation model and propensity score to decrease the variance of the IPS approach.
MF-DR~\cite{wang2019doubly} optimizes the following DR estimator of the ideal loss.
\begin{align}
&\drL (\predR; \matO)  \notag \\
&:= \frac{1}{mn} \SUM \left\{ \hat{\ell}_{u,i} + \iO \frac{ \ell ( \iR, \ipredR ) - \hat{\ell}_{u,i}} {\iP}  \right\} , 
\label{eq:dr_loss}
\end{align}

where $\hat{\ell}_{u,i}$ is the \textit{imputation model}, which estimates $ \ell ( \iR, \ipredR ) $.
As discussed in \cite{wang2019doubly}, this estimator satisfies the unbiasedness with the true propensity (i.e., $ \mE [ \drL ] = \idealL $) and often achieves a tighter estimation error bound compared to IPS.
MF-DR obtains the final prediction ($\predR$) and the imputation model simultaneously via a joint learning procedure.

However, the proposed joint learning algorithm still requires a pre-estimated propensity score~\cite{wang2019doubly}. 
Furthermore, the estimation performance of the DR estimator is significantly degraded when the error imputation model and the propensity model are both misspecified~\cite{dudik2011doubly}. 
In fact, in the empirical evaluations of~\cite{wang2019doubly}, MCAR test data are used to estimate the propensity score.
In the experiments, we validate the performance of the DR estimator in a more practical setting where only MNAR data are available (without any MCAR data).

\paragraph{\textbf{Causal Embeddings (CausE)}}
The work most closely related to ours is~\cite{bonner2018causal}, which proposed CausE, a domain adaptation-inspired method to address the selection bias in rating feedback. The loss function of CausE is given as follows: 
\begin{align}
& \causeL ( \predR_{MCAR}, \predR_{MNAR}; \matO_{MCAR}, \matO_{MNAR} ) \notag \\
& \ := \naiveL ( \predR_{MCAR}; \matO_{MCAR} ) + \naiveL ( \predR_{MNAR}; \matO_{MNAR} )  \notag \\
& \quad + \beta \cdot \Omega_{domain} (\predR_{MCAR}, \predR_{MNAR}),  \label{eq:cause_loss}
\end{align}
where $\beta \ge 0$ is a trade-off hyper-parameter. 
Two prediction matrices ($\predR_{MCAR} $ and $\predR_{MNAR}$) are for the MCAR and MNAR data, respectively, where $\predR_{MNAR}$ is used to make the final predictions.
The last term represents the \textit{regularizer between tasks} and penalizes the divergence between the predictions for the MCAR and MNAR data.

This method empirically outperforms propensity-based methods in a binary rating feedback setting~\cite{bonner2018causal}.
A problem is that this algorithm, by its design, requires some MCAR data, which are generally unavailable. 
Moreover, it uses the idea of domain adaptation in a heuristic manner by myopically adding $\Omega_{domain}(\cdot)$; there is no theoretical guarantee for its loss function.
Therefore, we explore an algorithm that is preferable to CausE in the following aspects: 
(i) Our algorithm does not use any MCAR data during training, and thus is feasible in a realistic situation having no MCAR training data,
(ii) Our proposed method is theoretically refined in the sense that it minimizes the propensity-independent upper bound of the ideal loss function.

\section{Proposed Framework and Algorithm} \label{sec:method}
This section derives a propensity-independent generalization error bound for the ideal loss and an algorithm to minimize the bound using only MNAR data.

\subsection{Theoretical Bound}
First, we define a discrepancy measure to quantify the similarity between two different propensity matrices.
\begin{definition}(Propensity Matrix Divergence (PMD))
Let $\mH$ be a set of real-valued predicted rating matrices and $ \predR \in \mH $ be a specific prediction. The PMD between any two given propensity matrices $\matP$ and $\matP^{\prime}$ is defined as follows:
\begin{align*}
    \truePMD := \sup_{ \predR^{\prime} \in \mH } \ \{ \shortL (\predR, \predR^{\prime}; \matP ) - \shortL ( \predR, \predR^{\prime}; \matP^{\prime} ) \},
\end{align*}
where $\shortL (\predR, \predR^{\prime}; \matP)  :=  \mathbb{E}_{\matO \sim \matP} [ \widehat{\mathcal{L}}^{\ell}_{naive} (\predR, \predR^{\prime}; \matO) ] =  \frac{1}{M} \sum_{(u,i) \in \D} \iP \cdot \ell (\ipredR, \hat{R}^{\prime}_{u,i}) . $
\end{definition}

Note that the PMD is well defined because $ \shortPMD (\matP, \matP) = 0, \forall \matP \in \mathcal{P} $, and it satisfies nonnegativity and subadditivity.
Moreover, it is independent of the true rating matrices and can be calculated for any given pair of propensity matrices without true rating information.

We now use $\shortPMD$ and derive a propensity-independent generalization error bound for the ideal loss function.\footnote{Note that the theoretical bounds of all existing studies depend on the cardinality of the hypothesis space $|\mathcal{H}|$~\cite{schnabel2016recommendations,wang2019doubly}.
This is unrealistic, as $|\mathcal{H}|$ is infinite in almost all cases (as the parameter space is continuous). Under a realistic setting with infinite $\mathcal{H}$, therefore, the existing analysis becomes invalid. Thus, we use the Rademacher complexity to consider a realistic situation with infinite hypothesis spaces.}
\begin{theorem}(Propensity Independent Generalization Error Bound) 
Suppose two observation matrices with MCAR and MNAR mechanisms ($\mOmcar \sim \mPmcar$ and $\mOmnar \sim \mPmnar$) are given. 
For any prediction matrix $\predR \in \mH $ and for any $\delta \in (0, 1)$, the following inequality holds with a probability of at least $1 - \delta$:
\begin{align}
    &\idealL  (\predR) \notag \\
    & \le \naiveL (\predR; \mOmnar )  +  \shortPMD (\mOmcar, \mOmnar ) \notag \\ 
    & \quad + 2 L ( 3 \rad (\mH) + 2 \prad (\mH) ) + 3 \Delta \sqrt{\frac{\log (6/\delta)}{2M}} , \label{eq:theoretical_bound}
\end{align}
where $L$ is the Lipschitz constant of the loss function and $\rad (\mH)$ is the Rademacher complexity of $\mH$, which is defined in Appendix~\ref{sec:thm1}. See also Appendix~\ref{sec:thm1} for the proof. 
\label{thm1}
\end{theorem}

The theoretical bound consists of the following four factors: 
(i) Naive loss on MNAR data,
(ii) Empirical PMD,
(iii) The complexity of the hypothesis class, 
(iv) Confidence term that depends on $\delta$.
Note that (i) and (ii) can be optimized with only observable data, and (iii) can be controlled by adding a regularization term, as described in the next section. 
Morevoer (iv) converges to zero as $M$ increases.
We will empirically show that the bound is informative in the sense that optimizing (i) and (ii) results in a desired value of the ideal loss function.

\begin{algorithm}[t]
\caption{Domain Adversarial Matrix Factorization}
\label{alg:damf}
\begin{algorithmic}[1]
\REQUIRE MNAR observation matrix $ \mOmnar $; set of user-item pairs $\mathcal{D}$; trade-off parameter $\beta$; regularization parameter $\lambda$; batch size $B$; number of steps $k$
\ENSURE learnt prediction matrix $\predR = \matU \matV^{\top}$
\STATE Randomly initialize $\matU$, $\matV$, $\matU^{\prime}$, and $\matV^{\prime}$
\REPEAT
    \STATE Sample mini-batch data of size $B$ from $\mOmnar$
    \FOR{$1, \ldots, k$}
        \STATE Update $\matU$ and $\matV$ by gradient descent according to \Eqref{eq:approx_objective} with fixed $\predR^{*} = \matU^{\prime} ( \matV^{\prime} )^{\top}$
    \ENDFOR
    \STATE Uniformly sample user--item pairs of size $T$ from $\D$ to construct $\mOmcar$
    \FOR{$1, \ldots, k$}
        \STATE Update $\matU^{\prime}$ and $\matV^{\prime}$ by gradient ascent according to \Eqref{eq:approx_pmd} with fixed $\predR = \matU \matV^{\top}$
    \ENDFOR
\UNTIL{convergence;}
\end{algorithmic}
\end{algorithm}

\subsection{Algorithm}
Here, we describe the proposed algorithm.
Building on the theoretical bound derived in \Thmref{thm1}, we aim at minimizing the following objective:
\begin{align}
    & \min_{\predR \in \mH} 
     \underbrace{\naiveL ( \predR; \mOmnar )}_{\text{naive loss on MNAR training data}} \notag \\
    &\quad + \beta \cdot \underbrace{\shortPMD (\mOmcar, \mOmnar )}_{\text{discrepancy}}
    + \lambda \cdot \underbrace{\Omega (\predR) }_{\text{regularization}}, \label{eq:objective}
\end{align}
where $\beta \ge 0$ is the trade-off hyper-parameter between the naive loss and discrepancy measure. $\Omega (\cdot)$ is a regularization function that penalizes the complexity of $\predR$, and $\lambda \ge 0$ is the regularization hyper-parameter. 
In addition, we can obtain $\mOmcar$ by uniformly sampling unlabeled user--item pairs from $\D$.
The objective in \Eqref{eq:objective} builds on the two controllable terms of the theoretical bound in \Eqref{eq:theoretical_bound}.
Thus it is independent of the propensity score, and we need \textbfit{not} estimate the propensity score to optimize this objective. 
Note that we add the regularization term to our objective to control the complexity of the hypothesis class $\mH$, as it is impossible to minimize the Rademacher complexity directly.

To solve \Eqref{eq:objective}, by definition, we empirically approximate $\shortPMD$ by solving
\begin{align}
    \max_{ \predR^{\prime} \in \mH } \ \naiveL ( \predR, \predR^{\prime}; \mOmcar ) - \naiveL ( \predR, \predR^{\prime}; \mOmnar ). \label{eq:approx_pmd}
\end{align}
This optimization corresponds to accurately estimating PMD from observable data.

Given the above details, the optimization in \Eqref{eq:objective} is reduced to:
\begin{align}
    & \min_{\predR \in \mH} 
     \underbrace{\frac{1}{M} \sum_{(u,i): \iO = 1 }  \ell (\iR, \hat{R}_{u,i})}_{\text{empirical loss on MNAR training data}}  \notag \\
    & \quad +  \beta \cdot \underbrace{\{  \naiveL ( \predR, \predR^{*}; \mOmcar ) - \naiveL ( \predR, \predR^{*}; \mOmnar )  \}}_{\text{approximated PMD between MCAR and MNAR}} \notag \\
    & \quad +  \lambda \cdot \underbrace{\Omega (\predR) }_{\text{regularization}}, \label{eq:approx_objective}
\end{align}
where $\predR^{*}$ is the solution of \Eqref{eq:approx_pmd}.

Algorithm~\ref{alg:damf} describes the detailed procedure of our proposed DAMF, which is based on the matrix factorization (MF) model, where a prediction matrix is defined as the product of user--item latent factors $\matU \in \mathbb{R}^{m \times d}$ and $\matV \in \mathbb{R}^{n \times d}$. Given that the predictions are modelled by MF, a possible regularization is $\Omega (\predR) := \|\matU\|_{2}^{2}+\|\matV\|_{2}^{2}$, where $\|\cdot\|_2$ is the L2 norm.

\begin{table}[ht]
\centering
\caption{Rating Prediction Performance}
\vspace{2mm}
\scalebox{0.9}{
\begin{tabular}{cc|cc}
    \toprule
    Datasets & Methods && MSE ($\pm$ StdDev) \\ \midrule \midrule
    \multirow{7}{*}{Yahoo! R3} 
    & MF && 1.7343 ($\pm$ 0.0309)  \\ 
    & MF-IPS && 1.7320 ($\pm$ 0.0311)   \\ 
    & MF-DR && 1.7445 ($\pm$ 0.0189)  \\ 
    & DAMF (ours) && \textbf{1.2787 ($\pm$ 0.0126)}  \\\cmidrule{2-4}
    & CausE && 1.7390 ($\pm$ 0.0283) \\ 
    & MF-IPS (true) && 1.1281 ($\pm$ 0.0161)  \\ 
    & MF-DR (true) && 1.0435 ($\pm$ 0.0166)  \\ \midrule
    \multirow{7}{*}{Coat} 
    & MF && 1.2166 ($\pm$ 0.0007)   \\ 
    & MF-IPS && 1.2044 ($\pm$ 0.0012)  \\ 
    & MF-DR && 1.2108 ($\pm$ 0.0016)   \\ 
    & DAMF (ours) && \textbf{1.1371 ($\pm$ 0.0025)}  \\ \cmidrule{2-4}
    & CausE && 1.2801 ($\pm$ 0.0043)  \\ 
    & MF-IPS (true) && 1.0675 ($\pm$ 0.0043)  \\
    & MF-DR (true) && 1.0760 ($\pm$ 0.0037)  \\
\bottomrule
\end{tabular}}
\label{tab:results}
\vskip 0.05in
\raggedright
\fontsize{9pt}{9pt}\selectfont \textit{Note}: 
The bold font indicates the best performance in each metric and dataset among methods using only MNAR data.
\end{table}

\begin{table*}[ht]
\centering
\caption{Rating Prediction Performance of MF-IPS and MF-DR with Different Propensity Estimators}
\vspace{2mm}
\scalebox{0.9}{
\begin{tabular}{cc|ccccc}
    \toprule
    Datasets & Methods &&  \textit{user propensity} & \textit{item propensity} & \textit{user-item propensity} & \textit{1BitMC} \\ 
    \midrule \midrule
    %
    \multirow{2}{*}{Yahoo! R3} 
    & MF-IPS && 1.8216 (+61.5\%) & 1.8083 (+60.3\%) & 1.8714 (+65.9\%) & 1.7343 (+53.8\%)  \\ 
    & MF-DR && 1.7445 (+67.2\%) & 1.9074 (+82.8\%) & 1.8815 (+80.4\%) & 1.8953 (+81.6\%)   \\  \midrule
    \multirow{2}{*}{Coat}
    & MF-IPS && 1.2171 (+14.0\%) & 1.2161 (+13.9\%) & 1.2044 (+12.8\%) & 1.2087 (+13.2\%)  \\ 
    & MF-DR && 1.2141 (+12.8\%) & 1.2108 (+12.5\%) & 1.2181 (+13.2\%) & 1.2195 (+13.3\%) \\ 
\bottomrule
\end{tabular}}
\label{tab:pscore_results}
\vskip 0.05in
\raggedright
\fontsize{9pt}{9pt}\selectfont \textit{Note}: 
Performances relative to the true propensity score are in parentheses.
The result suggests that the performances of MF-IPS and MF-DR using only MNAR data are significantly worse than the performances of those using the true propensity.
\end{table*}
\begin{table}[ht]
\centering
\caption{Ranking Performance on \textbf{Yahoo!R3}}
\vspace{2mm}
\scalebox{0.8}{
\begin{tabular}{c|cccc}
    \toprule
    Methods  && NDCG@5 ($\pm$ StdDev) & Recall@5 ($\pm$ StdDev) \\ \midrule \midrule
    MF &&  0.7645 ($\pm$ 0.0008) & 0.5616 ($\pm$ 0.0006) \\ 
    MF-IPS && 0.7645 ($\pm$ 0.0008) & 0.5616 ($\pm$ 0.0006)  \\ 
    MF-DR && 0.7589 ($\pm$ 0.0007) & 0.5602 ($\pm$ 0.0005) \\ 
    DAMF (ours) && \textbf{0.7904 ($\pm$ 0.0011)} & \textbf{0.5737 ($\pm$ 0.0004)}  \\\cmidrule{1-4}
    CausE && 0.7645 ($\pm$ 0.0007) & 0.5617 ($\pm$ 0.0006) \\ 
    MF-IPS (true) && 0.7454 ($\pm$ 0.0021) & 0.5565 ($\pm$ 0.0011)  \\ 
    MF-DR (true) && 0.7629 ($\pm$ 0.0032) & 0.5624 ($\pm$ 0.0015) \\
\bottomrule
\end{tabular}}
\label{tab:ranking_results_yahoo}
\vspace{-5mm}
\end{table}

\begin{table}[ht]
\centering
\caption{Ranking Performance on \textbf{Coat}}
\vspace{2mm}
\scalebox{0.8}{
\begin{tabular}{c|cccc}
    \toprule
    Methods  && NDCG@5 ($\pm$ StdDev) & Recall@5 ($\pm$ StdDev) \\ \midrule \midrule
    MF && 0.6617 ($\pm$ 0.0029) & 0.3858 ($\pm$ 0.0008)  \\ 
    MF-IPS && 0.6717 ($\pm$ 0.0026) & 0.3844 ($\pm$ 0.001)   \\ 
    MF-DR && 0.6789 ($\pm$ 0.0018) & 0.3887 ($\pm$ 0.0016)   \\ 
    DAMF (ours) && \textbf{0.6847 ($\pm$ 0.0034)} & \textbf{0.3901 ($\pm$ 0.0011)}  \\ \cmidrule{1-4}
    CausE && 0.6332 ($\pm$ 0.0039) & 0.3752 ($\pm$ 0.0013)  \\ 
    MF-IPS (true) && 0.6772 ($\pm$ 0.002) & 0.3906 ($\pm$ 0.0009)  \\
    MF-DR (true) && 0.6758 ($\pm$ 0.0014) & 0.3900 ($\pm$ 0.0012)  \\
\bottomrule
\end{tabular}}
\label{tab:ranking_results_coat}
\vskip 0.05in
\raggedright
\fontsize{8.5pt}{8.5pt}\selectfont \textit{Note}: 
The bold font indicates the best performance in each metric and dataset among methods using only MNAR data.
\end{table}

\section{Experiments} \label{sec:experiment}
This section empirically evaluates the proposed method on public real-world datasets. Appendix~\ref{app:experiments} describes detailed experimental settings such as dataset description, evaluation metrics, and hyper-parameter tuning.

\subsection{Experimental Setups}
\paragraph{\textbf{Datasets.}}
We use Yahoo! R3,\footnote{http://webscope.sandbox.yahoo.com/} and Coat.\footnote{https://www.cs.cornell.edu/\textasciitilde schnabts/mnar/}.
These datasets are originally divided into the MNAR training set and the MCAR test set.  We randomly selected 10\% of the original training set as the validation set, which is used to perform hyper-parameter tuning.

\paragraph{\textbf{Baseline Methods and Propensity Estimators.}}
We compare the following methods with our proposed method: 
(i) \textbf{MF~\cite{koren2009matrix}}, which optimizes its model parameters by minimizing the naive loss in \Eqref{eq:naive_loss} and does not depend on the propensity score.
(ii) \textbf{MF-IPS~\cite{schnabel2016recommendations}}, which optimizes its model parameters by minimizing the IPS loss in \Eqref{eq:ips_loss}.
(iii) \textbf{MF-DR~\cite{wang2019doubly}}, which optimizes its model parameters by minimizing the DR loss in \Eqref{eq:dr_loss}.
(iv) \textbf{CausE~\cite{bonner2018causal}}, which minimizes the sum of the naive loss and the domain regularization term $\Omega_{domain}(\cdot)$, which measures the divergence between the MNAR and MCAR data. To calculate its loss function, we sample 5\% of the MCAR test data. 
MCAR data are in general unavailable in real-world settings, and thus we report the results of CausE simply as a reference.

For MF-IPS and MF-DR, we use \textit{user propensity}, \textit{item propensity}, \textit{user--item propensity}, and \textit{1BitMC}~\cite{ma2019missing} as the variants of the propensity estimator and report the results with the best estimator for each dataset. 
Note that these four propensity estimators are usable in real-world recommender systems because they use only MNAR data.
These estimators are formally defined in Appendix~\ref{app:experiments}.

We also report the results of MF-IPS and MF-DR with the \textit{true propensity score}.
We follow previous studies~\cite{schnabel2016recommendations,wang2019doubly} and calculate this true propensity using 5\% of the MCAR test data.
Note that the true propensity is unavailable in real-world situations, as it requires MCAR explicit feedback.
Therefore, we report the results of MF-IPS and MF-DR with the true propensity score as a reference.

\subsection{Results and Discussion} \label{sec:results}
Here, we report and discuss the experimental results with respect to four research questions (RQ1-RQ4). We run the experiments 10 times with different initializations and report the averaged results in Tables 1-4.

\paragraph{\textbf{\textit{RQ1. How well does the proposed algorithm perform on the rating prediction task?}}}
We evaluate and compare the prediction performance of DAMF and the baselines on Yahoo! R3 and Coat.
Table~\ref{tab:results} provides the MSEs and their standard deviations (StdDev).
For Yahoo! R3, DAMF reveals the best rating prediction performance among the methods using only MNAR data.
Specifically, it outperforms MF-IPS by 26.3\% and MF-DR by 26.7\% in terms of MSE.
For Coat, DAMF outperforms MF-IPS by 5.59\% and MF-DR by 6.08\%, even though the distributional shift of Coat is significantly smaller than that of Yahoo! R3.

\paragraph{\textbf{\textit{RQ2. How do propensity-based methods perform with different propensity estimators?}}} 
Here we evaluate the sensitivity of MF-IPS and MF-DR to the choice of the propensity estimator.

First, consistent with previous studies~\cite{schnabel2016recommendations,wang2019doubly}, Table~\ref{tab:results} shows that MF-IPS and MF-DR with the true propensity exhibit better rating prediction performance compared to the other methods.
However, as shown in \Tabref{tab:pscore_results}, they perform poorly with other propensity estimators, including \textit{1BitMC}, and especially for Yahoo! R3.
In some cases, they even underperform MF, which is based on the naive loss.
Therefore, although propensity-based methods are potentially high-performing with the true propensity, their prediction performances are highly sensitive to the choice of propensity estimator and negatively affected by the propensity estimation bias.
These observations empirically demonstrate the \textbfit{propensity contradiction} of the previous methods.

\paragraph{\textbf{\textit{RQ3. How well does the proposed algorithm perform on the ranking task?}}}
Next, we test the ranking performance of DAMF on Yahoo! R3 and Coat.
Tables~\ref{tab:ranking_results_yahoo} and~\ref{tab:ranking_results_coat} describe the ranking metrics computed on the MCAR test data.

For Yahoo! R3, DAMF demonstrates the best ranking performance among all methods, including the methods with the true propensity score. 
It outperforms MF-IPS by 3.39\% and MF-DR by 4.15\%, MF-IPS (true) by 6.04\%, and MF-DR (true) by 3.61\% in NDCG.
The results suggest that DAMF is powerful and useful for improving the recommendation quality and user experience with only biased rating data.
Moreover, DAMF is the best in Coat, improving MF-IPS by 3.48\%, MF-DR by 1.94\%, MF-IPS (true) by 1.10\%, and MF-DR (true) by 1.32\% in NDCG.
These results demonstrate that the proposed method works satisfactorily in the ranking task even when the dataset size is small and the degree of bias is not significant.
Note that it is reasonable to assume that there exists a user--item pair with zero observed probability in Coat.
This is because the training set was collected via workers' self-selection, and male workers did not provide the ratings of women's coats and vice versa.
Thus, this result suggests the stability and adaptability of the proposed method to the data with a user--item pair with $\iP=0$.
In contrast, the performance of the propensity-based methods on Coat is not theoretically grounded, as the training and test sets may not overlap.

\begin{figure}[t]
\begin{center}
    \begin{tabular}{c}
        \begin{minipage}{0.48\hsize}
            \begin{center}
                \includegraphics[clip, width=4cm]{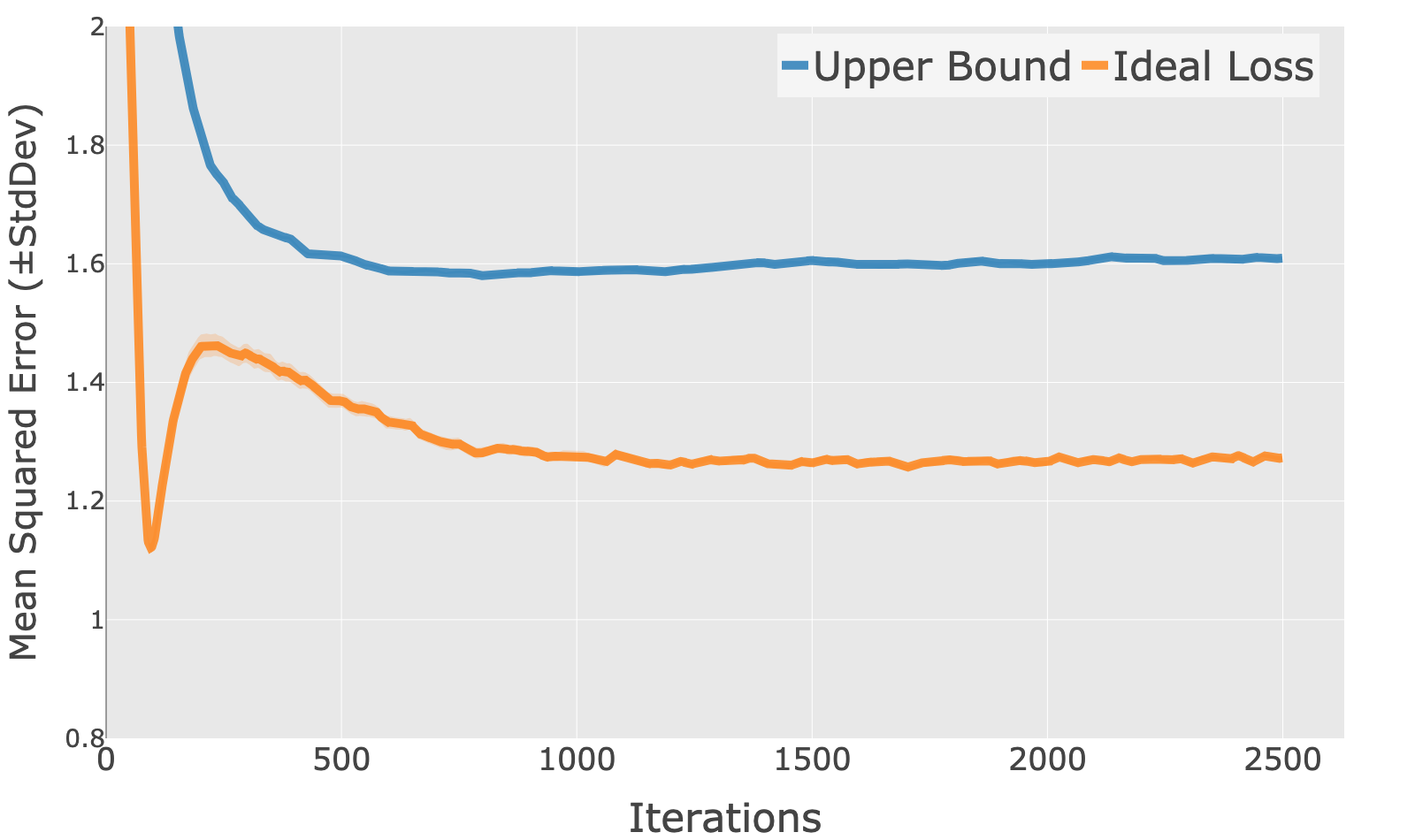}
                \vskip 0.03in
                (a) Yahoo! R3
            \end{center}
        \end{minipage}
        
        \begin{minipage}{0.48\hsize}
            \begin{center}
                \includegraphics[clip, width=4cm]{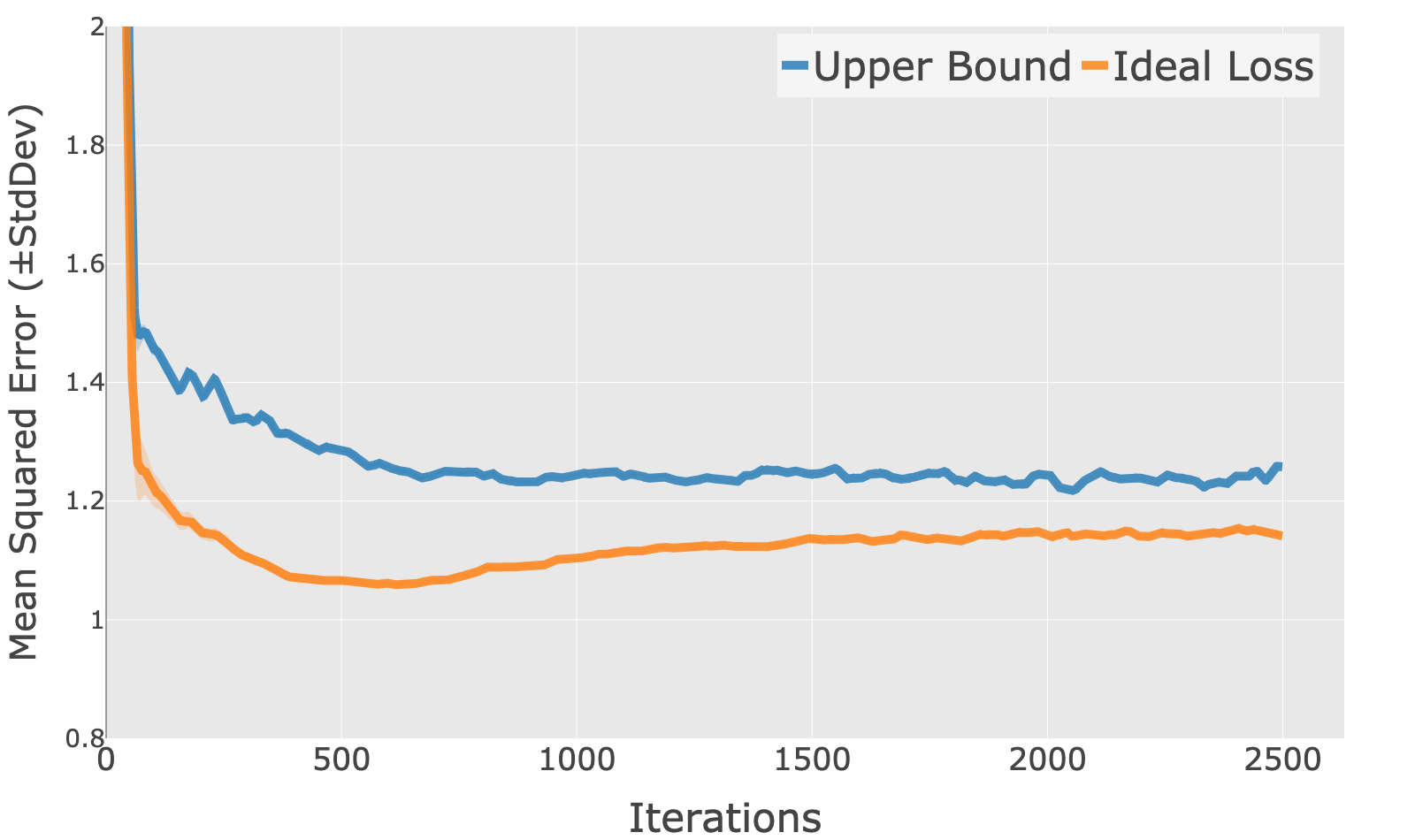}
                \vskip 0.03in
                (b) Coat 
            \end{center}
        \end{minipage}
    \end{tabular}
\end{center}
\caption{Theoretical Upper Bound and Ideal Loss Function during DAMF Training.}
\label{fig:upper_bound_curves}
\end{figure}

\paragraph{\textbf{\textit{RQ.4 Does adversarial learning really minimize ideal loss?}}}
Finally, we investigate the validity of minimizing the propensity-independent upper bound in \Eqref{eq:theoretical_bound} as a solution to minimize the ideal loss function in \Eqref{eq:ideal_loss} using only MNAR data.  
\Figref{fig:upper_bound_curves} depicts the sum of the first two terms of the upper bound (blue) and ideal loss (orange) during the training of DAMF on Yahoo! R3 and Coat.\footnote{We omit the constant terms that do not matter here in the upper bound, and thus the ideal loss may be smaller.}
First, it is evident from the figures that our adversarial learning procedure effectively minimizes the upper bound of the ideal loss during training.
Moreover, the figures suggest that the ideal loss function is also minimized well by our method.
These results demonstrate that minimizing the theoretical upper bound in \Eqref{eq:theoretical_bound} is a valid approach for improving the recommendation quality on the MCAR test set.

\section{Conclusion} \label{sec:conclusion}
We explored the problem of offline recommender learning from \textbfit{only} MNAR explicit rating feedback. To this end, we derived the \textbfit{propensity independent generalization error bound} of the loss function and proposed an algorithm to minimize the bound via adversarial learning. Our algorithm enables an accurate rating prediction without estimating the propensity score, thereby resolving the \textbfit{propensity contradiction} of the existing literature. Extensive experiments demonstrate the advantages of the proposed framework in terms of rating prediction and ranking measures when true propensities are inaccessible.

Our work represents an initial attempt for offline recommender learning via adversarial training, and there are still many problems to study in this direction. 
For example, our theory leads to a propensity-independent upper bound of the (ideal) rating prediction loss.
It would be valuable to construct a corresponding theory for the ranking task, such as showing the propensity-independent upper bound of the additive ranking metrics.
Moreover, testing our method in an online experiment will support the reliability of our framework. We leave the investigation of these directions for future studies.

\newpage
\clearpage
\appendix

\section{Omitted Proofs} \label{sec:thm1}

Throughout this section, we use the following \textit{Rademacher complexity}~\cite{bartlett2002rademacher,mohri2018foundations}, which captures
the complexity of a class of functions by measuring its ability to correlate with random noise~\cite{kuroki2019unsupervised}.
\begin{definition}(Rademacher complexity) 
Let $\mH$ be any set of real-valued matrices.
Given i.i.d samples with observed ratings $\{ (u,i,\iR) \mid \iO = 1, (u,i) \in \D \}$, the Rademacher complexity of $\mH$ is
\begin{align*}
    \rad (\mH) := 
    \mE_{\matO \sim \matP} \ \mE_{\boldsymbol{\xi}} \left[  \sup_{ \predR \in \mH } \frac{1}{M} \sum_{(u,i): \iO = 1} \xi_{u,i} \ipredR \right],
\end{align*}
where $\boldsymbol{\xi} = ( \xi_1, \ldots \xi_M ) $ denotes a set of independent uniform random variables taking values in $\{+1, -1\}$.
\end{definition}

\subsection{Uniform Derivation Bound}
\begin{lemma} (Rademacher Generalization Bound; A modified version of Theorem 3.3 in \cite{mohri2018foundations})
Let $\mathcal{F} = \{ f : \D \rightarrow [0, \Delta] \}$ be a class of bounded functions where $\Delta > 0$ is a positive constant and $ \{ (u,i,\iR) \mid \iO = 1, (u,i) \in \D \} $ be any i.i.d. sample drawn from $\matP$ of size $M$.
Then, for any $\delta \in (0,1)$, the following inequality holds with probability of at least $1-\delta$
\begin{align}
    & \sup_{f \in \mathcal{F}} \ \Bigl| \underbrace{\frac{1}{M} \sum_{(u,i): \iO=1} f (u,i)}_{(a)} -  \underbrace{\frac{1}{mn} \SUM \iP \cdot f (u,i)}_{(b)} \Bigr| \notag  \\
    & \qquad \qquad \le 2 \rad ( \mathcal{F} ) + \Delta \sqrt{ \frac{\log (2/\delta)}{2M}},
    \label{eq:rad_bound}
\end{align}
where $(a)$ is an empirical mean of a function ($f$), and $(b)$ is its expectation over $\matP$. 
\label{lem:rad_bound}
\end{lemma}

\subsection{Deviation Bound of PMD} \label{sec:lem1}
The true propensity matrices ($\matP$ and $\matP^{\prime}$) are unknown, and it is necessary to estimate the PMD using their realizations ($\matO$ and $\matO^{\prime}$). 
The following lemma shows the deviation bound of PMD.
\begin{lemma}
Suppose any pair of propensity matrices ($\matP$ and $\matP^{\prime}$) and their realizations ($\matO$ and $\matO^{\prime}$) are given. 
Accordingly, for any $\delta \in (0,1)$ and $ \mH $, the following inequality holds with a probability of at least $1 - \delta$:
\begin{align*}
  & \left| \truePMD - \shortPMD \left(\matO, \matO^{\prime} \right) \right|  \\
  & \ \le 4L \left( \rad (\mH ) + \prad (\mH) \right) + \Delta \sqrt{ \frac{2\log (4/\delta) }{M} }.
\end{align*}
\label{lem1}
\end{lemma}
\begin{proof}
For any given real-valued prediction matrix $\predR$, we have
\begin{align}
    & \Bigl| \shortPMD (\matP, \matP^{\prime} ) - \shortPMD (\matO, \matO^{\prime} ) \Bigr| \notag \\
    & = \Bigl| \sup_{\predR^{\prime} \in \mH } \ \{ \shortL ( \predR, \predR^{\prime} ; \matP  ) - \shortL ( \predR, \predR^{\prime} ; \matP^{\prime}  ) \} \notag \\
    & \quad - \sup_{\predR^{\prime} \in \mH } \ \{ \shortL ( \predR, \predR^{\prime} ; \matO  ) - \shortL ( \predR, \predR^{\prime} ; \matO^{\prime}  ) \} \Bigr|  \notag \\
    & \le \sup_{\predR^{\prime} \in \mH } \ \Bigl| \{ \shortL ( \predR, \predR^{\prime} ; \matP  ) - \shortL ( \predR, \predR^{\prime} ; \matP^{\prime}  )  \}  \notag \\
    & \quad - \{ \shortL ( \predR, \predR^{\prime} ; \matO ) - \shortL ( \predR, \predR^{\prime} ; \matO^{\prime}  )  \} \Bigr| \notag \\
    & = \sup_{\predR^{\prime} \in \mH } \ \Bigl| \{ \shortL ( \predR, \predR^{\prime} ; \matP  ) - \shortL ( \predR, \predR^{\prime} ; \matO  ) \}  \notag \\
    & \quad - \{ \shortL ( \predR, \predR^{\prime} ; \matP^{\prime}  ) - \shortL ( \predR, \predR^{\prime} ; \matO^{\prime}  )  \} \Bigr| \notag \\
    & = \sup_{\predR, \predR^{\prime} \in \mH } \ \Bigl| \{ \shortL ( \predR, \predR^{\prime} ; \matP  ) - \shortL ( \predR, \predR^{\prime} ; \matO  )  \}  \notag \\ 
    & \quad  - \{ \shortL ( \predR, \predR^{\prime} ; \matP^{\prime}  ) - \shortL ( \predR, \predR^{\prime} ; \matO^{\prime}  )  \} \Bigr| \notag \\
    & \le \sup_{\predR, \predR^{\prime} \in \mH } \ \Bigl| \shortL ( \predR, \predR^{\prime} ; \matP ) - \shortL ( \predR, \predR^{\prime} ; \matO  )  \Bigr|   \notag \\
    & \quad + \sup_{\predR, \predR^{\prime} \in \mH } \ \Bigl| \shortL ( \predR, \predR^{\prime} ; \matP^{\prime}  ) - \shortL ( \predR, \predR^{\prime} ; \matO^{\prime}  )  \Bigr|.  \label{eq13}
\end{align}
The deviations in the last line can be bounded by using \Lemref{lem:rad_bound}, and the following inequalities hold with a probability of at least $1 - \delta / 2$.
\begin{align}
  &\sup_{\predR^{\prime} \in \mH } \ \Bigl| \shortL ( \predR, \predR^{\prime} ; \matP ) - \shortL ( \predR, \predR^{\prime} ; \matO  )  \Bigr| \notag \\
  &\qquad \le  2  \rad (\mH^{\prime}) + \Delta \sqrt{ \frac{\log (4/\delta)}{2M}},  \label{eq14} \\
  &\sup_{\predR^{\prime} \in \mH } \ \Bigl| \shortL ( \predR, \predR^{\prime} ; \matP^{\prime} ) - \shortL ( \predR, \predR^{\prime} ; \matO^{\prime}  )  \Bigr| \notag \\
  &\qquad \le 2  \prad (\mH^{\prime}) + \Delta \sqrt{ \frac{\log (4/\delta)}{2M}}.  \label{eq15}
\end{align}
where we regard $\mH^{\prime} \coloneqq \{ (u,i) \rightarrow \ell (\ipredR, \ipredR^{\prime} ) \mid \predR, \predR^{\prime} \in \mH \}$ as $\mathcal{F}$ in \Lemref{lem:rad_bound}.
Then, we have
\begin{align}
    & \sup_{\predR, \predR^{\prime} \in \mH } \ \Bigl| \shortL ( \predR, \predR^{\prime} ; \matP ) - \shortL ( \predR, \predR^{\prime} ; \matO  )  \Bigr| \notag \\
    &\qquad \le  4L \prad (\mH) + \Delta \sqrt{ \frac{\log (4/\delta)}{2M}},  \label{eq14b} \\
    & \sup_{\predR, \predR^{\prime} \in \mH } \ \Bigl| \shortL ( \predR, \predR^{\prime} ; \matP^{\prime} ) - \shortL ( \predR, \predR^{\prime} ; \matO^{\prime}  )  \Bigr| \notag \\
    &\qquad \le 4L \rad (\mH ) + \Delta \sqrt{ \frac{\log (4/\delta)}{2M}}.  \label{eq15b}
\end{align}
where $\rad (\mH^{\prime}) \le  2L \rad (\mH) $ by using the result of Corollary 5 of~\cite{mansour2009domain}.
Finally, combining \Eqref{eq13}, \Eqref{eq14b}, and \Eqref{eq15b} with the union bound completes the proof.
\end{proof}

\subsection{Proof of Additional Lemmas}
Here, we state the generalization error bound when using the naive loss.
\begin{lemma}(Generalization Error Bound)
An MCAR-observation matrix $\matO \sim \matP$ and any matrix as predictions $ \predR \in \mH $ are given. 
Then, for any $\delta \in (0, 1)$, the following inequality holds with a probability of at least $1 - \delta$:
\begin{align}
    \shortL ( \predR; \matP) 
    & \le \naiveL ( \predR ; \mO) \notag \\
    &\qquad + 2L \rad (\mH) + \Delta \sqrt{ \frac{\log (2 / \delta)}{2M} } . \label{eq:generalization_bound_mcar}
\end{align}
\begin{proof}
By using \Lemref{lem:rad_bound}, we have
\begin{align*}
    &\sup_{\predR \in \mH} \Bigl| \shortL (\predR ; \matP) -  \naiveL ( \predR ; \matO) \Bigr| \notag \\
    &\qquad \le 2 \rad (\ell \circ \mH) + \Delta \sqrt{ \frac{\log (2 / \delta)}{2M}}.
\end{align*}
with a probability of at least $1-\delta$ for any $\delta \in (0,1)$.
We regard $ \ell \circ \mH =  \{ \D \rightarrow \ell (\iR, \ipredR ) \mid \predR \in \mH \} $ as $\mathcal{F}$ in \Lemref{lem:rad_bound}.
Then, by using the Talagrand's lemma (Lemma 5.7 of \cite{mohri2018foundations}), we have
\begin{align*}
    \rad (\ell \circ \mH) \le L \rad (\mH) .
\end{align*}
as $\ell$ is $L$-lipschitz.
\end{proof}
\label{lem2}
\end{lemma}

\begin{lemma}
For any given predicted rating matrix $ \widehat{ \trueR} \in \mH $ and two propensity matrices ($\matP$ and $\matP^{\prime}$), the following inequality holds
\begin{align*}
    \shortL ( \predR; \matP ) \le \shortL ( \predR ; \matP^{\prime} ) + \truePMD.
\end{align*}
\begin{proof}
By the definition of PMD, we have
\begin{align*}
    \shortL ( \predR; \matP ) 
    & =  \shortL ( \predR; \matP^{\prime} ) - \shortL ( \predR; \matP^{\prime} ) + \shortL ( \predR; \matP ) \\
    & \le \shortL ( \predR; \matP^{\prime} ) + \sup_{ \predR^{\prime} \in \mH } \{ \shortL ( \predR, \matP) - \shortL ( \predR; \matP^{\prime} ) \} \\
    & = \shortL ( \predR ; \matP^{\prime} ) + \truePMD.
\end{align*}
\end{proof}
\label{lem3}
\end{lemma}
    
\subsection{Proof of \Thmref{thm1}}
\begin{proof}
First, we obtain the following inequality by replacing $ \matP $ and $ \matP^{\prime} $ for $ \mPmcar $ and $ \mPmnar $ in \Lemref{lem3}. 
\begin{align}
    \idealL (\predR) \le & \widehat{\mathcal{L}}^{\ell} ( \predR; \mPmnar)\notag \\
    &\qquad + \shortPMD (\mPmcar, \mPmnar), \label{eq16}
\end{align}
where $ \idealL (\predR) = \shortL (\predR; \mPmcar) $ by definition.
Then, from \Lemref{lem1} and \Lemref{lem2}, the following inequalities hold with a probability of at least $1 - 2\delta / 3$ and $1 - \delta / 3$, respectively.
\begin{align}
    & \shortL (\predR; \mPmnar ) \le \naiveL ( \predR; \mOmnar )  \notag \\
    & \quad + 2L \rad (\mH) + \Delta \sqrt{ \frac{\log (6 /\delta)}{2M} }, \label{eq17} \\
    & \Bigl| \shortPMD  (\mPmcar, \mPmnar) - \shortPMD  (\mOmcar, \mOmnar ) \Bigr| \notag \\
    & \quad \le 4L (\rad (\mH) + \prad (\mH) ) + 2 \Delta \sqrt{\frac{\log (6 / \delta) }{2M}}. \label{eq18}
\end{align}
Combining \Eqref{eq16}, \Eqref{eq17}, and \Eqref{eq18} with the union bound completes the proof.
\end{proof}

\section{Experiment Details} \label{app:experiments}
\begin{figure*}[ht]
\begin{center}
    \begin{tabular}{c}
        \begin{minipage}{0.4\hsize}
            \begin{center}
                Yahoo! R3 ($\textit{KL-div}=0.470$)
                \includegraphics[clip, width=6cm]{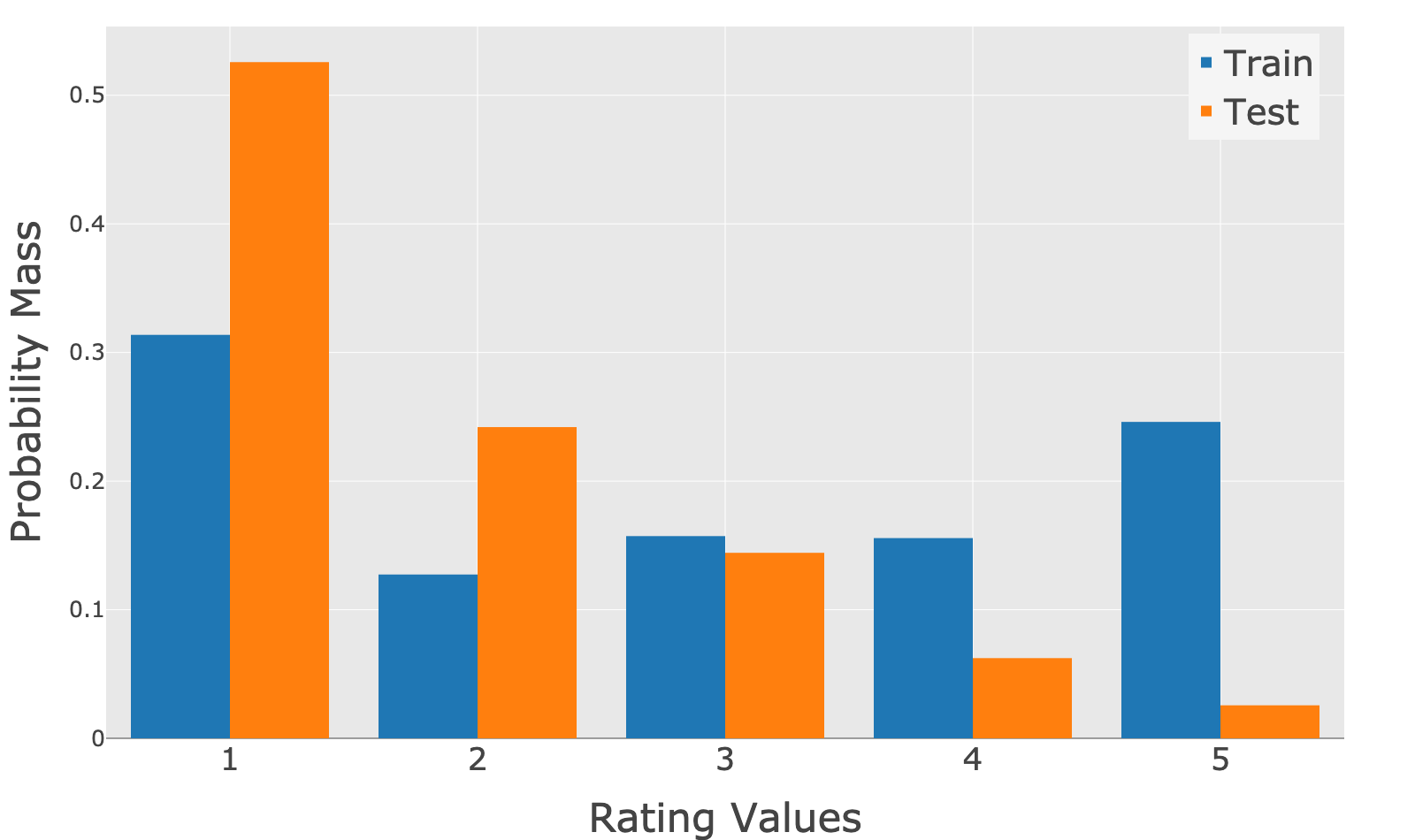}
            \end{center}
        \end{minipage}
        
        \begin{minipage}{0.4\hsize}
            \begin{center}
                Coat ($\textit{KL-div}=0.049$)
                \includegraphics[clip, width=6cm]{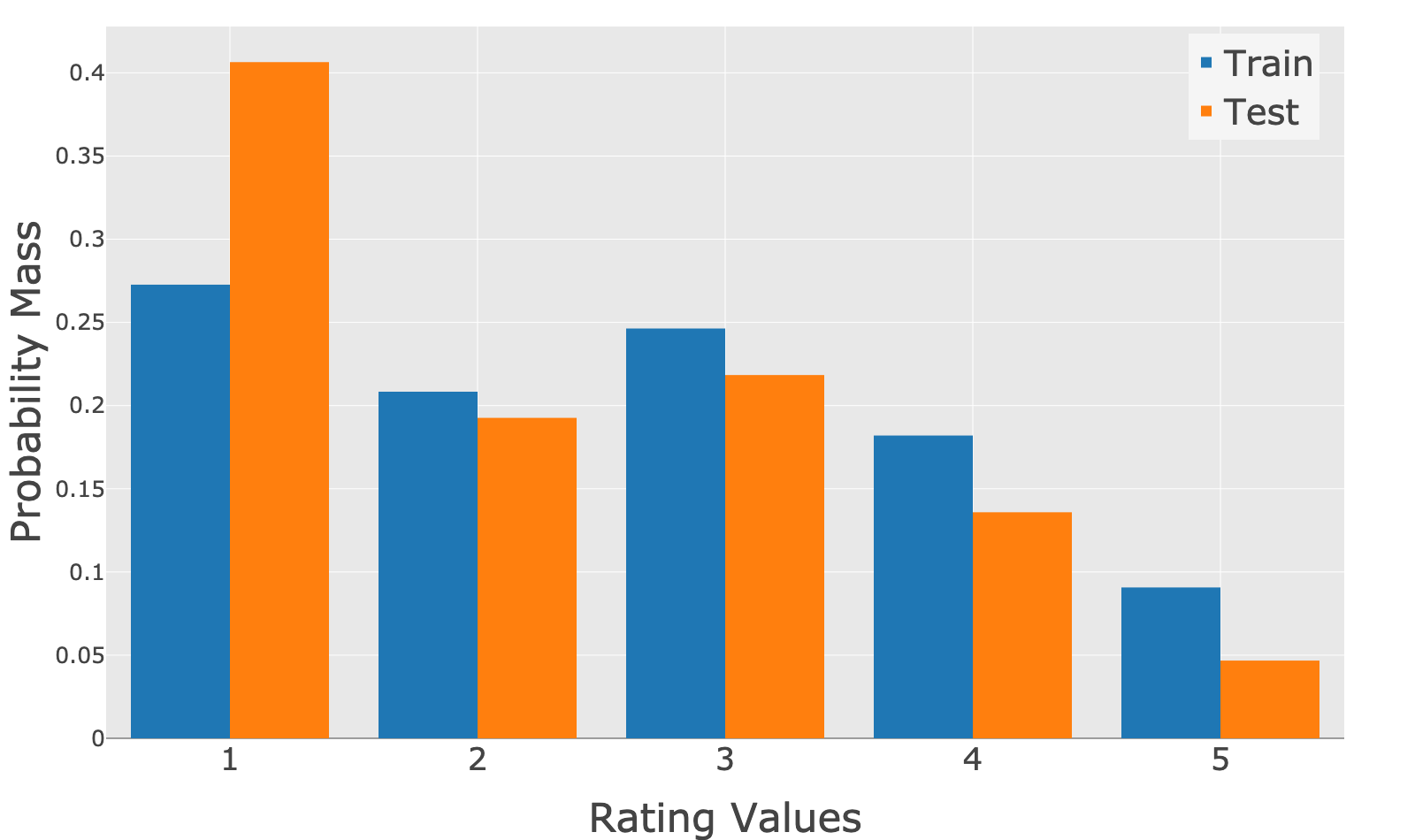}
            \end{center}
        \end{minipage}
    \end{tabular}
\end{center}
\caption{Rating Distributions of Training and Test Sets of Yahoo! R3 and Coat.}
\vskip 0.05in
\raggedright
\fontsize{9pt}{9pt}\selectfont \textit{Note}: 
The rating distributions are significantly different between the training and test sets for both datasets. 
Note that \textit{KL-div} is the Kullback–Leibler divergence of the rating distributions between training and test sets.
Therefore, the distributional shift of Yahoo! R3 dataset is relatively large compared to that of the Coat dataset.
\label{fig:rating_dists}
\end{figure*}

\paragraph{\textbf{Datasets.}}
We used the following real-world datasets. 
\begin{itemize}
    \item Yahoo! R3\footnote{http://webscope.sandbox.yahoo.com/}: It contains five-star user-song ratings. The training data contain approximately 300,000 MNAR ratings from 15,400 users for 1,000 songs, and the test data were collected by a subset of users to rate 10 randomly sampled songs.
    \item Coat\footnote{https://www.cs.cornell.edu/\textasciitilde schnabts/mnar/}: It contains five-star user-coat ratings from 290 Amazon Mechanical Turk workers on an inventory of 300 coats. The training data contain 6,500 MNAR ratings collected via self-selection by Turk workers. Conversely, the test data were collected by asking Turk workers to rate 16 randomly selected coats.
\end{itemize}
Note here that Yahoo! R3 and Coat are the only publicity available real-world datasets that contain the test sets with MCAR ratings.
\Figref{fig:rating_dists} shows the training and test rating distributions of Yahoo! R3 and Coat. The figures clearly show the distributional shifts between the training and test sets.

\paragraph{\textbf{Definition of Propensity Estimators.}}

For MF-IPS and MF-DR, we use \textit{user propensity}, \textit{item propensity}, \textit{user--item propensity}, and \textit{1BitMC}~\cite{ma2019missing} as the variants of the propensity estimator and report the results with the best estimator for each dataset. 
Note that these four propensity estimators are usable in real-world recommender systems because they use only MNAR data.
These four estimators are defined as follows:
\begin{align*}
& \textit{user propensity} \ : \  \widehat{P}_{u, *} := \frac{\sum_{i \in \mathcal{I}} O_{u,i}}{\max_{u \in U} \sum_{i \in \mathcal{I}} O_{u,i}}, \\
& \textit{item propensity} \ : \  \widehat{P}_{*, i} := \frac{\sum_{u \in \mathcal{U}} O_{u,i}}{\max_{i \in I} \sum_{u \in \mathcal{U}} O_{u,i}}, \\
& \textit{user-item propensity} \ : \  \widehat{P}_{u, i} := \widehat{P}_{u, *} \cdot \widehat{P}_{*, i}, \\
& \textit{1BitMC} \ :  \\
& \hat{\Gamma}_{u, i} := \argmin_{\Gamma \in \mathcal{F}_{\tau, \gamma}} 
\sum_{(u,i) \in \D} \{ \iO \log ( \sigma (\Gamma_{u, i}) )\\
&\ \ \ \ \ \ \ \ \ \ \ +  (1 - \iO) \log (1 - \sigma (\Gamma_{u, i})) \} 
\end{align*}
where $ \mathcal{F}_{\tau, \gamma} \coloneqq \{ \Gamma \in \mathbb{R}^{m\times n} \ | \ || \Gamma ||_* \le \tau \sqrt{mn}, \ || \Gamma ||_{\mathrm{max}} \le \gamma  \} $, $ || \cdot ||_* $ ($\tau, \gamma > 0$) denotes the nuclear norm, and $ ||\cdot ||_{\mathrm{max}} $ is the entry-wise max norm.
$\sigma(\cdot)$ is the sigmoid function, and $ \hat{P}_{u, i} \coloneqq \sigma (\hat{\Gamma}_{u, i}) $.
We use the implementation provided by Ma et al.~\cite{ma2019missing} for 1BitMC. Note that 1BitMC is a sophisticated version of the logistic regression to estimate the propensity score described in Schnabel et al.~\cite{schnabel2016recommendations}.

\paragraph{\textbf{hyper-parameter Tuning.}}
Table~\ref{tab:hyperparam_search_space} shows the hyper-parameter search space used in the experiments.
We tune the dimensions of the latent factors $d$ in the range of $\{5, 10, \ldots, 40 \}$ and the L2-regularization hyper-parameter $\lambda$ in the range of $[ 10^{-4}, 1]$ for all methods. For DAMF and CausE, the trade-off hyper-parameter $\beta$ is also tuned in the range of $[10^{-2}, 1]$.  We identify the best values for these hyper-parameters by running an adaptive hyper-parameter tuning procedure implemented in \textit{Optuna}~\cite{akiba2019optuna} on a validation set.

\begin{table}[h]
\centering
\caption{Hyper-parameter Search Space}
\vspace{2mm}
\scalebox{0.75}{
\begin{tabular}{cc|ccccc} 
\toprule
    Datasets & Methods && $d$ & $\lambda$ & $\beta$  \\ 
    \midrule \midrule
    \multirow{4}{*}{Yahoo! R3} & MF-IPS && \multirow{4}{*}{$\{5, 10, \ldots, 40 \}$} & \multirow{4}{*}{$[ 10^{-4}, 1]$} & -   \\ 
    & MF-DR && &  &     \\ 
    & CausE && &  & \multirow{2}{*}{$[10^{-2}, 1]$} \\ 
    & DAMF (ours) && &  &   \\  
    \midrule
    \multirow{4}{*}{Coat} & MF-IPS && \multirow{4}{*}{$\{5, 10, \ldots, 40 \}$} & \multirow{4}{*}{$[ 10^{-8}, 1]$} &  -  \\ 
    & MF-DR && &  &   \\ 
    & CausE && &  & $ [10^{-10}, 1] $ \\ 
    & DAMF (ours) && &  & $ [10^{-2}, 1] $   \\ 
\bottomrule
\end{tabular}}
\vskip 0.05in
\raggedright
\fontsize{9pt}{9pt}\selectfont \textit{Note}: 
$d$ is the dimension of the latent factors, $\lambda$ is the hyperparameter for the L2-regularization, and $\beta$ is the trade-off hyperparameter.
We also used the Adam optimizer with $0.01$ as its initial learning rate for all methods.
\label{tab:hyperparam_search_space}
\end{table}
\paragraph{\textbf{Performance Measures.}}
The following defines the performance measures used in the experiments.

\begin{itemize}
\item \textbf{MSE} evaluates how far the predicted ratings are away from the true rating and is defined as
    \begin{align*}
        \textit{MSE} := \frac{1}{|\D_{\mathrm{te}}|} \sum_{(u,i) \in \D_{\mathrm{te}}} ( \iR - \ipredR )^2,
    \end{align*}
    where $\D_{\mathrm{te}}$ is the user--item pairs in the test set.
\item \textbf{Normalized discounted cumulative gain (NDCG)} measures the ranking quality and is defined as:
    \begin{align*}
         & \textit{DCG@K}  := \frac{1}{m} \sum_{(u,i)} \frac{2^{(\iR-1)} \cdot \mathbb{I} \{ rank(u,i) \le K \}}{\log_2 (rank(u,i) + 1) }.
    \end{align*}
    where $\mathbb{I} \{\cdot\}$ is the indicator function, $rank(u,i)$ is a ranking of $i$ for $u$ induced by $\predR$.
    Then, $\textit{NDCG@K} := \textit{DCG@K}/\textit{IDCG@K},$ where IDCG@K is the maximum possible DCG@K.
\item \textbf{Recall} evaluates how many relevant items are selected and is defined as
    $$ \textit{Recall@K} := \frac{1}{m} \sum_{(u,i)} \frac{\iR \cdot \mathbb{I} \{ rank(u,i) \le K \}}{\sum_{i \in [n]} \iR}.$$
\end{itemize}

\section{Related Work} \label{sec:related_work}
This section summarizes the related literature.

\paragraph{\textit{\textbf{Offline Recommender Learning from MNAR Feedback}}:} 
To address the selection bias of MNAR explicit feedback, propensity-based methods have been explored~\cite{liang2016causal,schnabel2016recommendations,wang2019doubly}. 
Among these, the most basic method is the IPS estimation, which was originally established in causal inference~\cite{imbens2015causal,rosenbaum1983central,rubin1974estimating,saito2020counterfactual}. 
This method provides an unbiased estimator of the true metric of interest by weighting each data point using the inverse of its propensity. 
The rating predictor based on the IPS estimator empirically outperformed both the naive MF~\cite{koren2009matrix} and probabilistic generative models ~\cite{hernandez2014probabilistic}. 
IPS can reasonably remove the bias of naive methods, however, their performance depends on the quality of the propensity score estimation~\cite{saito2020unbiased,saito2020asymmetric,saito2020pairwise,saito2020dual}. 
Specifically, it is challenging to ensure the performance of the propensity estimation in real-world recommendations, as users are independent in selecting which items to rate, and one cannot control the missing mechanism~\cite{marlin2009collaborative}. 
In addition to the simple IPS estimator, Wang et al.~\cite{wang2019doubly} proposed the doubly robust (DR) variant to decrease the variance of the propensity-weighting approach. 
The DR estimator utilizes both the error imputation model and the propensity model, and theoretically improves the bias and estimation error bound compared to the IPS counterpart. 
However, the proposed joint learning algorithm still requires pre-estimated propensity scores~\cite{wang2019doubly}. 
Furthermore, the estimation performance of the DR estimator is significantly degraded when error imputation and propensity models are both misspecified~\cite{saito2019doubly,saito2020doubly,saito2020open,dudik2011doubly,wang2019doubly,kiyohara2022doubly,kallus2021optimal,saito2019doubly}. 
In the empirical evaluation of the propensity-based methods, MCAR test data were used to estimate the propensity score~\cite{schnabel2016recommendations,wang2019doubly}. 
However, MCAR data are unavailable in most practical situations, as gathering a sufficient amount of MCAR data requires more time and cost requirements for the annotation process~\cite{gilotte2018offline,saito2020open,saito2022off,saito2021counterfactual,saito2021evaluating}.

Currently, there are two studies that address issues related to conventional propensity-based methods.
Bonner and Vasile~\cite{bonner2018causal} proposed CausE, a domain adaptation inspired method that introduces a regularizer term on the discrepancy between latent factors obtained from MCAR and MNAR data. 
However, this method requires some MCAR training data, which are generally unavailable. 
Moreover, there is no theoretical guarantee for the proposed loss function.
Therefore, our method is more desirable than CausE in that
(i) Our method does not use any MCAR data in its training process and is feasible in a realistic situation with no MCAR data. 
(ii) Our method is theoretically refined as it is designed to minimize the propensity-independent upper bound of the ideal loss function.

Next, Ma et al.~\cite{ma2019missing} proposed a propensity estimation method, 1BitMC, which does not require MCAR data.
The authors constructed the theoretical guarantee for the consistency of 1BitMC.
However, it presupposes the use of \textbf{inverse} propensity weighting to debias downstream recommenders; it cannot be used when there exists a user--item pair with zero observed probability. 
Furthermore, its experiments use only small-size recommendation datasets.
In addition, the performance of the recommendation methods with 1BitMC was evaluated using only prediction accuracy measures.
We address the limitations of Ma et al.~\cite{ma2019missing} as follows:
(i) Our proposed method and theory do not depend on the propensity score and are applicable to settings where there exists a user--item pair with zero observed probability.
(ii) Through comprehensive experiments on some moderately sized datasets, we demonstrate that our proposed method performs better than MF-IPS and MF-DR with 1BitMC for both rating prediction and ranking tasks.

\paragraph{\textit{\textbf{Unsupervised Domain Adaptation (UDA)}}:}
The aim of UDA is to train a predictor that works well on a target domain by using only labeled source data and unlabeled target data during training~\cite{kuroki2019unsupervised,saito2017asymmetric}. 
However, the major challenge of UDA is that the feature distributions and labeling functions can differ between the source and target domains. 
Thus, a predictor trained using only labeled source data does not generalize well to the target domain. 
Therefore, it is essential to measure the dissimilarity between the domains to achieve a desired performance in the target domain~\cite{kuroki2019unsupervised,lee2019domain}. 
Several discrepancy measures have been proposed to measure the difference in feature distributions between the source and target domains~\cite{ben2010theory,kuroki2019unsupervised,lee2019domain,zhang2019bridging}.
For example, $\mathcal{H}$-divergence and $\mathcal{H}\Delta\mathcal{H}$-divergence~\cite{ben2010theory,ben2007analysis} have been used to construct many prediction methods for UDA, such as DANN, ADDA, and MCD~\cite{ganin2015unsupervised,ganin2016domain,tzeng2017adversarial,saito2018maximum}. 
These methods are based on the adversarial learning framework and can be theoretically explained as minimizing empirical errors and discrepancy measures between the source and target domains. 
Note that the optimization of these methods does not depend on the propensity score. 
Thus, UDA is useful in constructing an effective recommender with biased rating feedback, given the absence of access to the true propensity score.

This study extended the idea of using a discrepancy measure to quantify the difference between two propensity score matrices and derive a propensity-independent generalization error bound, which is non-trivial given only the UDA literature, as it does not handle matrices with different distributions. Moreover, we provided an algorithm to optimize the upper bound of the ideal loss function via adversarial learning and matrix factorization.

\bibliographystyle{named}
\bibliography{main}

\begin{thebibliography}{}

\bibitem[\protect\citeauthoryear{Akiba \bgroup \em et al.\egroup
  }{2019}]{akiba2019optuna}
Takuya Akiba, Shotaro Sano, Toshihiko Yanase, Takeru Ohta, and Masanori Koyama.
\newblock Optuna: A next-generation hyperparameter optimization framework.
\newblock In {\em Proceedings of the 25th ACM SIGKDD International Conference
  on Knowledge Discovery \& Data Mining}, KDD '19, pages 2623--2631, 2019.

\bibitem[\protect\citeauthoryear{Bartlett and
  Mendelson}{2002}]{bartlett2002rademacher}
Peter~L Bartlett and Shahar Mendelson.
\newblock Rademacher and gaussian complexities: Risk bounds and structural
  results.
\newblock {\em Journal of Machine Learning Research}, 3(Nov):463--482, 2002.

\bibitem[\protect\citeauthoryear{Ben-David \bgroup \em et al.\egroup
  }{2007}]{ben2007analysis}
Shai Ben-David, John Blitzer, Koby Crammer, and Fernando Pereira.
\newblock Analysis of representations for domain adaptation.
\newblock In {\em Advances in neural information processing systems}, pages
  137--144, 2007.

\bibitem[\protect\citeauthoryear{Ben-David \bgroup \em et al.\egroup
  }{2010}]{ben2010theory}
Shai Ben-David, John Blitzer, Koby Crammer, Alex Kulesza, Fernando Pereira, and
  Jennifer~Wortman Vaughan.
\newblock A theory of learning from different domains.
\newblock {\em Machine learning}, 79(1-2):151--175, 2010.

\bibitem[\protect\citeauthoryear{Bonner and Vasile}{2018}]{bonner2018causal}
Stephen Bonner and Flavian Vasile.
\newblock Causal embeddings for recommendation.
\newblock In {\em Proceedings of the 12th ACM Conference on Recommender
  Systems}, RecSys '18, pages 104--112, New York, NY, USA, 2018. ACM.

\bibitem[\protect\citeauthoryear{Chen \bgroup \em et al.\egroup
  }{2020}]{chen2020bias}
Jiawei Chen, Hande Dong, Xiang Wang, Fuli Feng, Meng Wang, and Xiangnan He.
\newblock Bias and debias in recommender system: A survey and future
  directions.
\newblock {\em arXiv preprint arXiv:2010.03240}, 2020.

\bibitem[\protect\citeauthoryear{Dud{\'\i}k \bgroup \em et al.\egroup
  }{2011}]{dudik2011doubly}
Miroslav Dud{\'\i}k, John Langford, and Lihong Li.
\newblock Doubly robust policy evaluation and learning.
\newblock In {\em Proceedings of the 28th International Conference on
  International Conference on Machine Learning}, pages 1097--1104, 2011.

\bibitem[\protect\citeauthoryear{Ganin and
  Lempitsky}{2015}]{ganin2015unsupervised}
Yaroslav Ganin and Victor Lempitsky.
\newblock Unsupervised domain adaptation by backpropagation.
\newblock In {\em Proceedings of the 32nd International Conference on Machine
  Learning}, volume~37 of {\em Proceedings of Machine Learning Research}, pages
  1180--1189. PMLR, 2015.

\bibitem[\protect\citeauthoryear{Ganin \bgroup \em et al.\egroup
  }{2016}]{ganin2016domain}
Yaroslav Ganin, Evgeniya Ustinova, Hana Ajakan, Pascal Germain, Hugo
  Larochelle, Fran{\c{c}}ois Laviolette, Mario Marchand, and Victor Lempitsky.
\newblock Domain-adversarial training of neural networks.
\newblock {\em The Journal of Machine Learning Research}, 17(1):2096--2030,
  2016.

\bibitem[\protect\citeauthoryear{Gilotte \bgroup \em et al.\egroup
  }{2018}]{gilotte2018offline}
Alexandre Gilotte, Cl{\'e}ment Calauz{\`e}nes, Thomas Nedelec, Alexandre
  Abraham, and Simon Doll{\'e}.
\newblock Offline a/b testing for recommender systems.
\newblock In {\em Proceedings of the Eleventh ACM International Conference on
  Web Search and Data Mining}, pages 198--206. ACM, 2018.

\bibitem[\protect\citeauthoryear{Hern{\'a}ndez-Lobato \bgroup \em et al.\egroup
  }{2014}]{hernandez2014probabilistic}
Jos{\'e}~Miguel Hern{\'a}ndez-Lobato, Neil Houlsby, and Zoubin Ghahramani.
\newblock Probabilistic matrix factorization with non-random missing data.
\newblock In {\em International Conference on Machine Learning}, pages
  1512--1520, 2014.

\bibitem[\protect\citeauthoryear{Imbens and Rubin}{2015}]{imbens2015causal}
Guido~W Imbens and Donald~B Rubin.
\newblock {\em Causal inference in statistics, social, and biomedical
  sciences}.
\newblock Cambridge University Press, 2015.

\bibitem[\protect\citeauthoryear{Joachims \bgroup \em et al.\egroup
  }{2017}]{joachims2017unbiased}
Thorsten Joachims, Adith Swaminathan, and Tobias Schnabel.
\newblock Unbiased learning-to-rank with biased feedback.
\newblock In {\em Proceedings of the Tenth ACM International Conference on Web
  Search and Data Mining}, pages 781--789. ACM, 2017.

\bibitem[\protect\citeauthoryear{Kallus \bgroup \em et al.\egroup
  }{2021}]{kallus2021optimal}
Nathan Kallus, Yuta Saito, and Masatoshi Uehara.
\newblock Optimal off-policy evaluation from multiple logging policies.
\newblock In {\em International Conference on Machine Learning}, pages
  5247--5256. PMLR, 2021.

\bibitem[\protect\citeauthoryear{Kiyohara \bgroup \em et al.\egroup
  }{2022}]{kiyohara2022doubly}
Haruka Kiyohara, Yuta Saito, Tatsuya Matsuhiro, Yusuke Narita, Nobuyuki
  Shimizu, and Yasuo Yamamoto.
\newblock Doubly robust off-policy evaluation for ranking policies under the
  cascade behavior model.
\newblock In {\em Proceedings of the 15th International Conference on Web
  Search and Data Mining}, pages 487--–497, 2022.

\bibitem[\protect\citeauthoryear{Koren \bgroup \em et al.\egroup
  }{2009}]{koren2009matrix}
Yehuda Koren, Robert Bell, and Chris Volinsky.
\newblock Matrix factorization techniques for recommender systems.
\newblock {\em Computer}, 42(8):30--37, 2009.

\bibitem[\protect\citeauthoryear{Kuroki \bgroup \em et al.\egroup
  }{2019}]{kuroki2019unsupervised}
Seiichi Kuroki, Nontawat Charoenphakdee, Han Bao, Junya Honda, Issei Sato, and
  Masashi Sugiyama.
\newblock Unsupervised domain adaptation based on source-guided discrepancy.
\newblock In {\em Proceedings of the AAAI Conference on Artificial
  Intelligence}, volume~33, pages 4122--4129, 2019.

\bibitem[\protect\citeauthoryear{Lee \bgroup \em et al.\egroup
  }{2019}]{lee2019domain}
Jongyeong Lee, Nontawat Charoenphakdee, Seiichi Kuroki, and Masashi Sugiyama.
\newblock Domain discrepancy measure using complex models in unsupervised
  domain adaptation.
\newblock {\em arXiv preprint arXiv:1901.10654}, 2019.

\bibitem[\protect\citeauthoryear{Liang \bgroup \em et al.\egroup
  }{2016}]{liang2016causal}
Dawen Liang, Laurent Charlin, and David~M Blei.
\newblock Causal inference for recommendation.
\newblock In {\em Causation: Foundation to Application, Workshop at UAI}, 2016.

\bibitem[\protect\citeauthoryear{Ma and Chen}{2019}]{ma2019missing}
Wei Ma and George~H Chen.
\newblock Missing not at random in matrix completion: The effectiveness of
  estimating missingness probabilities under a low nuclear norm assumption.
\newblock In {\em Advances in Neural Information Processing Systems}, pages
  14871--14880, 2019.

\bibitem[\protect\citeauthoryear{Mansour \bgroup \em et al.\egroup
  }{2009}]{mansour2009domain}
Yishay Mansour, Mehryar Mohri, and Afshin Rostamizadeh.
\newblock Domain adaptation: Learning bounds and algorithms.
\newblock In {\em Proceedings of The 22nd Annual Conference on Learning Theory
  (COLT 2009)}, Montr\'eal, Canada, 2009.

\bibitem[\protect\citeauthoryear{Marlin and
  Zemel}{2009}]{marlin2009collaborative}
Benjamin~M Marlin and Richard~S Zemel.
\newblock Collaborative prediction and ranking with non-random missing data.
\newblock In {\em Proceedings of the third ACM conference on Recommender
  systems}, pages 5--12. ACM, 2009.

\bibitem[\protect\citeauthoryear{Mnih and
  Salakhutdinov}{2008}]{mnih2008probabilistic}
Andriy Mnih and Ruslan~R Salakhutdinov.
\newblock Probabilistic matrix factorization.
\newblock In {\em Advances in neural information processing systems}, pages
  1257--1264, 2008.

\bibitem[\protect\citeauthoryear{Mohri \bgroup \em et al.\egroup
  }{2018}]{mohri2018foundations}
Mehryar Mohri, Afshin Rostamizadeh, and Ameet Talwalkar.
\newblock {\em Foundations of machine learning}.
\newblock MIT press, 2018.

\bibitem[\protect\citeauthoryear{Rosenbaum and
  Rubin}{1983}]{rosenbaum1983central}
Paul~R Rosenbaum and Donald~B Rubin.
\newblock The central role of the propensity score in observational studies for
  causal effects.
\newblock {\em Biometrika}, 70(1):41--55, 1983.

\bibitem[\protect\citeauthoryear{Rubin}{1974}]{rubin1974estimating}
Donald~B Rubin.
\newblock Estimating causal effects of treatments in randomized and
  nonrandomized studies.
\newblock {\em Journal of educational Psychology}, 66(5):688, 1974.

\bibitem[\protect\citeauthoryear{Saito and
  Joachims}{2021}]{saito2021counterfactual}
Yuta Saito and Thorsten Joachims.
\newblock Counterfactual learning and evaluation for recommender systems:
  Foundations, implementations, and recent advances.
\newblock In {\em Fifteenth ACM Conference on Recommender Systems}, pages
  828--830, 2021.

\bibitem[\protect\citeauthoryear{Saito and Joachims}{2022}]{saito2022off}
Yuta Saito and Thorsten Joachims.
\newblock Off-policy evaluation for large action spaces via embeddings.
\newblock {\em arXiv preprint arXiv:2202.06317}, 2022.

\bibitem[\protect\citeauthoryear{Saito and
  Yasui}{2020}]{saito2020counterfactual}
Yuta Saito and Shota Yasui.
\newblock Counterfactual cross-validation: Stable model selection procedure for
  causal inference models.
\newblock In {\em International Conference on Machine Learning}, pages
  8398--8407. PMLR, 2020.

\bibitem[\protect\citeauthoryear{Saito \bgroup \em et al.\egroup
  }{2017}]{saito2017asymmetric}
Kuniaki Saito, Yoshitaka Ushiku, and Tatsuya Harada.
\newblock Asymmetric tri-training for unsupervised domain adaptation.
\newblock In {\em Proceedings of the 34th International Conference on Machine
  Learning}, volume~70 of {\em Proceedings of Machine Learning Research}, pages
  2988--2997. PMLR, 2017.

\bibitem[\protect\citeauthoryear{Saito \bgroup \em et al.\egroup
  }{2018}]{saito2018maximum}
Kuniaki Saito, Kohei Watanabe, Yoshitaka Ushiku, and Tatsuya Harada.
\newblock Maximum classifier discrepancy for unsupervised domain adaptation.
\newblock In {\em Proceedings of the IEEE Conference on Computer Vision and
  Pattern Recognition}, pages 3723--3732, 2018.

\bibitem[\protect\citeauthoryear{Saito \bgroup \em et al.\egroup
  }{2019}]{saito2019doubly}
Yuta Saito, Hayato Sakata, and Kazuhide Nakata.
\newblock Doubly robust prediction and evaluation methods improve uplift
  modeling for observational data.
\newblock In {\em Proceedings of the 2019 SIAM International Conference on Data
  Mining}, pages 468--476. SIAM, 2019.

\bibitem[\protect\citeauthoryear{Saito \bgroup \em et al.\egroup
  }{2020a}]{saito2020open}
Yuta Saito, Shunsuke Aihara, Megumi Matsutani, and Yusuke Narita.
\newblock Open bandit dataset and pipeline: Towards realistic and reproducible
  off-policy evaluation.
\newblock {\em arXiv preprint arXiv:2008.07146}, 2020.

\bibitem[\protect\citeauthoryear{Saito \bgroup \em et al.\egroup
  }{2020b}]{saito2020dual}
Yuta Saito, Gota Morisihta, and Shota Yasui.
\newblock Dual learning algorithm for delayed conversions.
\newblock In {\em Proceedings of the 43rd International ACM SIGIR Conference on
  Research and Development in Information Retrieval}, pages 1849--1852, 2020.

\bibitem[\protect\citeauthoryear{Saito \bgroup \em et al.\egroup
  }{2020c}]{saito2020unbiased}
Yuta Saito, Yaginuma Suguru, Yuta Nishino, Hayato Sakata, and Nakata Kazuhide.
\newblock Unbiased recommender learning from missing-not-at-random implicit
  feedback.
\newblock In {\em Proceedings of the Thirteenth ACM International Conference on
  Web Search and Data Mining}. ACM, 2020.

\bibitem[\protect\citeauthoryear{Saito \bgroup \em et al.\egroup
  }{2021}]{saito2021evaluating}
Yuta Saito, Takuma Udagawa, Haruka Kiyohara, Kazuki Mogi, Yusuke Narita, and
  Kei Tateno.
\newblock Evaluating the robustness of off-policy evaluation.
\newblock In {\em Fifteenth ACM Conference on Recommender Systems}, pages
  114--123, 2021.

\bibitem[\protect\citeauthoryear{Saito}{2020a}]{saito2020asymmetric}
Yuta Saito.
\newblock Asymmetric tri-training for debiasing missing-not-at-random explicit
  feedback.
\newblock In {\em Proceedings of the 43rd International ACM SIGIR Conference on
  Research and Development in Information Retrieval}, pages 309--318, 2020.

\bibitem[\protect\citeauthoryear{Saito}{2020b}]{saito2020doubly}
Yuta Saito.
\newblock Doubly robust estimator for ranking metrics with post-click
  conversions.
\newblock In {\em Fourteenth ACM Conference on Recommender Systems}, pages
  92--100, 2020.

\bibitem[\protect\citeauthoryear{Saito}{2020c}]{saito2020pairwise}
Yuta Saito.
\newblock Unbiased pairwise learning from biased implicit feedback.
\newblock In {\em Proceedings of the 2020 ACM SIGIR on International Conference
  on Theory of Information Retrieval}, pages 5--12, 2020.

\bibitem[\protect\citeauthoryear{Schnabel \bgroup \em et al.\egroup
  }{2016}]{schnabel2016recommendations}
Tobias Schnabel, Adith Swaminathan, Ashudeep Singh, Navin Chandak, and Thorsten
  Joachims.
\newblock Recommendations as treatments: Debiasing learning and evaluation.
\newblock In {\em Proceedings of The 33rd International Conference on Machine
  Learning}, volume~48, pages 1670--1679. PMLR, 2016.

\bibitem[\protect\citeauthoryear{Steck}{2010}]{steck2010training}
Harald Steck.
\newblock Training and testing of recommender systems on data missing not at
  random.
\newblock In {\em Proceedings of the 16th ACM SIGKDD international conference
  on Knowledge discovery and data mining}, pages 713--722. ACM, 2010.

\bibitem[\protect\citeauthoryear{Tzeng \bgroup \em et al.\egroup
  }{2017}]{tzeng2017adversarial}
Eric Tzeng, Judy Hoffman, Kate Saenko, and Trevor Darrell.
\newblock Adversarial discriminative domain adaptation.
\newblock In {\em Proceedings of the IEEE Conference on Computer Vision and
  Pattern Recognition}, pages 7167--7176, 2017.

\bibitem[\protect\citeauthoryear{Wang \bgroup \em et al.\egroup
  }{2019}]{wang2019doubly}
Xiaojie Wang, Rui Zhang, Yu~Sun, and Jianzhong Qi.
\newblock Doubly robust joint learning for recommendation on data missing not
  at random.
\newblock In {\em International Conference on Machine Learning}, pages
  6638--6647, 2019.

\bibitem[\protect\citeauthoryear{Zhang \bgroup \em et al.\egroup
  }{2019}]{zhang2019bridging}
Yuchen Zhang, Tianle Liu, Mingsheng Long, and Michael Jordan.
\newblock Bridging theory and algorithm for domain adaptation.
\newblock In {\em International Conference on Machine Learning}, pages
  7404--7413, 2019.

\end{thebibliography}

\end{document}